\pdfoutput=1

\documentclass[11pt]{article}

\usepackage[final]{acl}

\usepackage{times}
\usepackage{latexsym}

\usepackage[T1]{fontenc}

\usepackage[utf8]{inputenc}

\usepackage{microtype}

\usepackage{inconsolata}

\usepackage{hyperref}       
\usepackage{url}            
\usepackage{booktabs}       
\usepackage{amsfonts}       
\usepackage{nicefrac}       
\usepackage{microtype}      
\usepackage{xcolor}         
\usepackage[inline]{enumitem}
\usepackage{graphicx}
\usepackage{dramatist}
\usepackage{dirtytalk}
\usepackage{multirow}
\usepackage{threeparttable}
\usepackage{subcaption}
\usepackage{caption}

\usepackage{algorithm}
\usepackage{algpseudocode}
\usepackage{amsmath}
\usepackage{amssymb}
\usepackage{booktabs}
\usepackage{multirow}
\colorlet{verylightgray}{lightgray!99!white}
\usepackage{listings}
\usepackage{tcolorbox}
\tcbuselibrary{skins,breakable}

\newcommand{\ourmethod}{PARSE}
\newcommand{\ourmethodlong}{Parameter Automated Refinement and Schema Extraction}
\newcommand{\ciemethod}{SCOPE}
\newcommand{\ciemethodlong}{Schema Compliant Organized Pattern Extractor}
\newcommand{\apmethod}{ARCHITECT}
\newcommand{\apmethodlong}{Automated Refinement and Conversion Handler for
Information Transformation and EnhanCemenT}

\title{\ourmethod: LLM Driven Schema Optimization for Reliable Entity Extraction}

\author{
  \textbf{Anubhav Shrimal\textsuperscript{*}},
  \textbf{Aryan Jain\textsuperscript{*}},
  \textbf{Soumyajit Chowdhury},
  \textbf{Promod Yenigalla}
\\
\\
  RBS Tech Sciences, Amazon
\\
  \small{
    \texttt{\{shrimaa, arynjn\}@amazon.com}
  }
\\
  \small{
    \textsuperscript{*}Equal contribution
  }
}

\begin{document}

\maketitle

\begin{abstract}
Structured information extraction from unstructured text is critical for emerging Software 3.0 systems where LLM agents autonomously interact with APIs and tools. Recent approaches apply large language models directly to extraction tasks using existing JSON schemas, often with constraint decoding or reinforcement learning approaches to ensure syntactic validity, but treat JSON schemas as static contracts designed for human developers, leading to suboptimal extraction performance, frequent hallucinations, and unreliable agent behavior when schemas contain ambiguous or incomplete specifications. We recognize that JSON schemas themselves are a form of natural language understanding contract that encodes rules, relationships, and expectations about data structure contracts that LLMs should be able to both interpret and systematically improve. Consequently, we develop PARSE (Parameter Automated Refinement and Schema Extraction), a novel system with two synergistic components: ARCHITECT, which autonomously optimizes JSON schemas for LLM consumption while maintaining backward compatibility through RELAY (an integrated code generation system), and SCOPE, which implements reflection-based extraction with combined static and LLM-based guardrails. We evaluate PARSE qualitatively and quantitatively on three datasets including Schema-Guided Dialogue (SGD), Structured Web Data Extraction (SWDE), and internal retail conversation data, and find that it achieves up to 64.7\% improvement in extraction accuracy on SWDE with combined framework improvements reaching 10\% across models, while reducing extraction errors by 92\% within the first retry and and maintaining practical latency.

\end{abstract}

\begin{figure*}[tbh]
  \centering
  \includegraphics[width=0.95\textwidth]{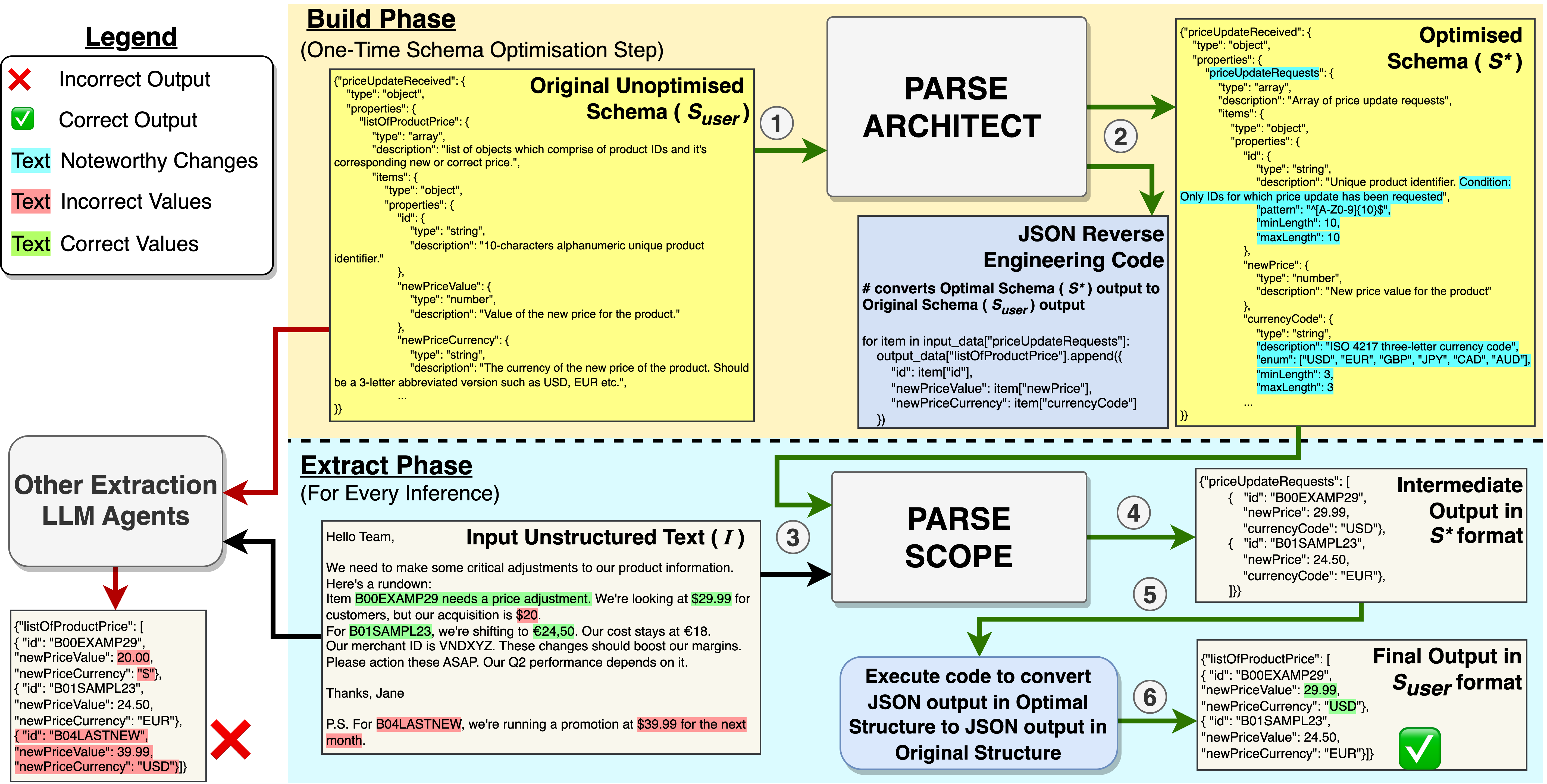}
  \caption{Overview of PARSE framework showing schema optimization and extraction pipeline. The system takes an unoptimized schema and input text (left), processes it through ARCHITECT for schema refinement and SCOPE for extraction (center), producing accurate structured output (right). The diagram highlights how PARSE outperforms other information extraction agents by implementing robust schema optimization and validation. Numbers indicate the sequential flow of operations.}
  \label{fig:parse_teaser_output}
\end{figure*}
\section{Introduction}
\label{sec:introduction}

The emergence of Software 3.0 represents a fundamental shift from static, form-based applications to dynamic systems where Large Language Model (LLM) agents autonomously interact with APIs and tools to accomplish complex tasks. In this new paradigm, reliable structured information extraction from unstructured text becomes mission-critical—agents must accurately parse natural language requests, extract precise parameters, and invoke the correct tools with valid arguments. Unlike traditional software where human developers handle data transformation complexity, LLM agents must perform this extraction reliably at scale, making the difference between a helpful assistant and a system that fails catastrophically in production.

What makes reliable structured extraction so challenging for LLM agents? Current approaches face a fundamental mismatch: JSON schemas that define expected output structures were designed as contracts between human developers and static systems, not as instructions for LLM agents. These schemas often contain ambiguous descriptions, incomplete validation rules, and structural choices optimized for human readability rather than machine comprehension. When LLM agents attempt extraction using these human-centric schemas, they struggle with unclear entity boundaries, conflicting requirements, and insufficient context about field relationships, leading to frequent hallucinations and schema non-adherence.

Existing work has focused primarily on forcing LLMs to conform to existing schema structures through constraint decoding, reinforcement learning, and self-correction mechanisms. However, these approaches treat schemas as immutable contracts rather than optimizing the structures themselves for LLM comprehension.

We observe that JSON schemas are themselves a form of natural language understanding contract—they encode rules, relationships, and expectations using descriptive text and validation logic that LLMs should be able to both interpret and systematically improve. Rather than viewing schemas as static artifacts, we can treat them as evolving interfaces optimized specifically for LLM consumption while maintaining backward compatibility with existing systems.

We present PARSE (Parameter Automated Refinement and Schema Extraction), a comprehensive system addressing both sides of the structured extraction challenge through two synergistic components. Together, these components create a virtuous cycle where schema optimization improves extraction performance, and extraction errors inform further schema refinement.

Our main contributions include:
\begin{enumerate*}[label=(\arabic*)]
\item \textbf{ARCHITECT (Automated Refinement and Conversion Handler for Information Transformation and EnhanCemenT):} A novel automated schema optimization framework that iteratively refines JSON schemas for LLM consumption by analyzing extraction performance, adding detailed entity descriptions and validation rules, and restructuring schemas for clearer representation while maintaining backward compatibility through RELAY, our automated transformation code generator. ARCHITECT includes RELAY (Reverse Engineering Layer for Automated Yoking), which automatically generates Python transformation code to maintain compatibility with original schema formats.
\item \textbf{SCOPE (Schema Compliant Organized Pattern Extractor):} A comprehensive reflection-based extraction framework that combines static and LLM-based guardrails to enable systematic error identification and correction, ensuring reliable structured information extraction through sophisticated validation mechanisms.
\end{enumerate*}
Our evaluation demonstrates that PARSE achieves up to 64.7\% improvement in extraction accuracy compared to state-of-the-art baselines on SWDE, with combined framework improvements reaching 10\% across SGD, SWDE, and internal retail conversation datasets, representing a paradigm shift toward treating structured extraction as a co-optimization problem between schema design and extraction mechanisms—enabling the reliable LLM agent systems that Software 3.0 applications demand.

\section{Related Work}
\label{sec:related_work}
Our work builds upon and extends several key research areas in information extraction, schema optimization, and LLM applications. This section provides a comprehensive review of relevant work across these domains. 

\noindent\textbf{LLM-based Structured Extraction and Schema Compliance:} Large Language Models have transformed information extraction from discriminative to generative approaches that produce structured outputs directly from unstructured text~\cite{5zhang-etal-2025-survey}. Two-stage frameworks combining general LLMs with domain-specific refinement show promise for complex extraction tasks~\cite{1zhang2024generalllmsinstructorsdomainspecific}, but current methods face significant limitations. LLMs struggle with complex schemas, exhibiting substantial performance gaps compared to traditional methods, particularly for nested entity recognition~\cite{4deng2024informationextractionlowresourcescenarios, 6han2024empiricalstudyinformationextraction}. Common error patterns include missing spans, incorrect types, and schema non-adherence, with GPT-4 showing an 11.97\% invalid response rate for complex extraction tasks~\cite{6han2024empiricalstudyinformationextraction}. Traditional JSON generation approaches have evolved from simple prompting to sophisticated constraint enforcement. Constraint decoding methods like Outlines guarantee schema compliance through grammar-guided generation but sometimes sacrifice output quality~\cite{14lu2025learninggeneratestructuredoutput, 16agarwal2025thinkinsidejsonreinforcement}. Reinforcement learning approaches achieve high valid JSON rates (98.7\% vs. 82.3\% baseline) by using schemas as training signals~\cite{16agarwal2025thinkinsidejsonreinforcement}, while the "Thought of Structure" paradigm shows substantial improvements (44.89\% gain) by encouraging structural reasoning before generation~\cite{15lu2025learninggeneratestructuredoutput}. However, these approaches focus on syntactic validity rather than optimizing schemas for LLM comprehension.

\noindent\textbf{Self-Correction and Agent Frameworks:} Self-correction techniques enhance LLM reliability through Chain-of-Thought reasoning, self-verification, and iterative refinement mechanisms~\cite{10kumar2025improvingreliabilityllmscombining}. The Chain of Self-Correction framework embeds correction as an inherent ability through iterative generation and verification~\cite{27gao2025embeddingselfcorrectioninherentability}. Runtime guardrail mechanisms like AgentSpec provide lightweight constraint enforcement through domain-specific languages~\cite{31wang2025agentspeccustomizableruntimeenforcement}, but existing methods primarily target general text generation rather than structured data extraction with strict schema adherence. LLM agents have emerged as critical frameworks for autonomous tool manipulation and API interaction~\cite{33luo2025largelanguagemodelagent}. The ReAct framework demonstrates synergized reasoning and acting through interleaved traces and actions~\cite{19yao2023reactsynergizingreasoningacting}, while systems like RestGPT enable direct RESTful API interaction~\cite{33luo2025largelanguagemodelagent}. Despite advances, agents face reliability challenges including high inference latency, output uncertainty, and security vulnerabilities~\cite{37liang2025llmpoweredaiagentsystems}, with "Agentic ROI" highlighting fundamental trade-offs between value and operational costs~\cite{34liu2025realbarrierllmagent}.

While substantial progress exists in LLM-based extraction, JSON schema optimization, and self-correction mechanisms, existing work treats these as independent problems. Current schema optimization focuses on syntactic validity rather than extraction performance, and self-correction mechanisms aren't designed for structured data adherence. Most critically, existing approaches assume static, human-designed schemas rather than exploring schema optimization for LLM consumption in Software 3.0 applications where LLM agents are primary consumers. Our work addresses this gap through an integrated approach combining automated schema optimization specifically for LLM agents with reflection-based guardrails tailored for structured extraction tasks.

\begin{figure}[tbh]
  \centering
  \includegraphics[width=\columnwidth]{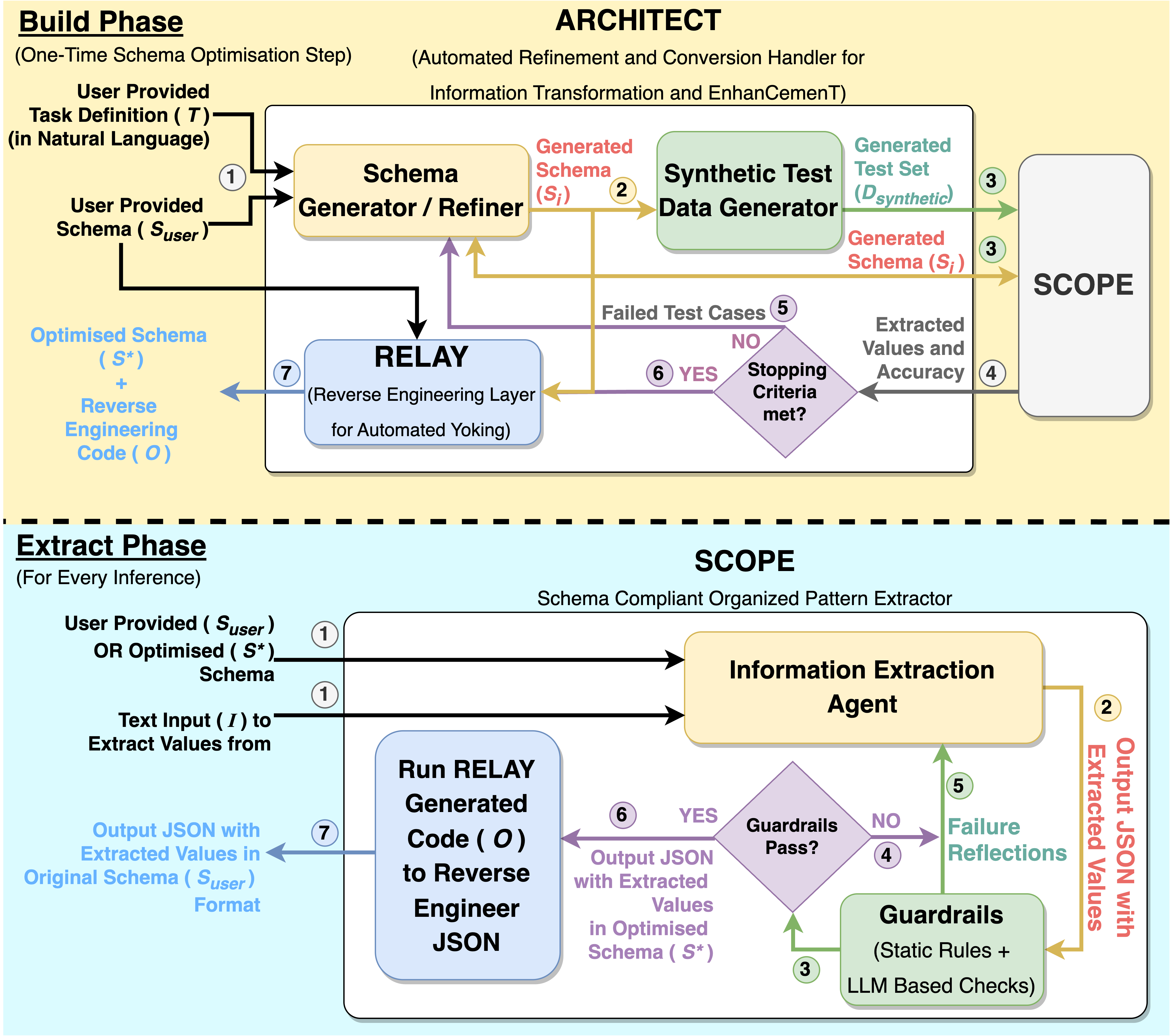}
  \caption{Detailed architecture of PARSE's two main components: ARCHITECT and SCOPE. The Build Phase (top) shows ARCHITECT's workflow for schema optimization through iterative refinement using synthetic test data generation and validation. The Extract Phase (bottom) illustrates SCOPE's extraction pipeline with built-in guardrails and reflection mechanisms for ensuring reliable structured output. Numbers indicate the sequential flow of operations.}
  \label{fig:parse_flow_diagram}
\end{figure}
\section{\ourmethod: \ourmethodlong}
\label{sec:approach}
PARSE addresses the challenge of reliable structured information extraction from unstructured text through a two-phase approach. Our key insight is that JSON schemas themselves represent natural language contracts that can be optimized for LLM consumption, creating a virtuous cycle where better schemas lead to improved extraction performance. PARSE consists of two primary components operating in distinct phases:
\begin{enumerate*}[label=(\arabic*)] 
    \item \textbf{Build Phase:} ARCHITECT optimizes JSON schemas, as a one time process, specifically for LLM agent consumption while maintaining backward compatibility through RELAY,
    \item \textbf{Extract Phase:} SCOPE performs reliable information extraction using reflection-based guardrails over the optimised schemas created and maps it back to original schema through RELAY transformation code generated in Build Phase.
\end{enumerate*}
Refer Figure~\ref{fig:parse_flow_diagram} for detailed flow.

\subsection{\apmethod:~\apmethodlong}
\textbf{Problem Formulation:} Let $S_{\text{user}}$ represent a user-provided JSON schema and $T$ represent a natural language task description. Current schemas are designed for human developers and static systems, leading to suboptimal performance when consumed by LLM agents. We formalize the schema optimization problem as:

$$S^* = \arg\max_{S'} \mathcal{P}(S', D_{\text{synthetic}}, M_{\text{llm}})$$

where $\mathcal{P}(S', D_{\text{synthetic}}, M_{\text{llm}})$ represents the extraction performance of LLM model $M_{\text{llm}}$ using schema $S'$ on synthetic validation data $D_{\text{synthetic}}$.

\noindent\textbf{Schema Generation and Refinement:} ARCHITECT begins by analyzing the original schema structure and generating an optimized version that enhances LLM comprehension through improved descriptions, structural clarity, and additional validation rules. The schema optimization process follows an iterative refinement procedure:

$$S_{i+1} = \text{Refine}(S_i, \mathcal{E}(S_i, D_{\text{synthetic}}), T)$$

where $S_i$ is the schema at iteration $i$, $\mathcal{E}(\cdot)$ computes extraction errors on test data $D_{\text{synthetic}}$ and could be done via SCOPE or any other information extraction agent, and $\text{Refine}(\cdot)$ represents the LLM-based schema improvement function. The optimization process terminates when extraction accuracy reaches a threshold $\tau$ or maximum iterations $K$ are reached.

\noindent\textbf{Synthetic Test Data Generation:} To validate schema improvements, ARCHITECT generates synthetic test cases using both the current schema and a seed dataset of real examples to ensure the synthetic data reflects realistic extraction scenarios:
$$D_{\text{synthetic}} = \text{Generate}(S_i, T, D_{\text{seed}}, n_{\text{samples}})$$

where $S_i$ is the current schema iteration, $T$ is the task description, $n_{\text{samples}}$ controls the diversity of generated validation examples, and The seed dataset $D_{\text{seed}}$ provides crucial context for generating realistic adversarial examples: 
$D_{\text{seed}} = \{(x_1, y_1), (x_2, y_2), \ldots, (x_k, y_k)\}$,
where each $x_j$ is unstructured text and $y_j$ is the corresponding ground truth extraction following the schema structure.

\noindent\textbf{RELAY (Reverse Engineering Layer for Automated Yoking):} To maintain compatibility with existing systems, ARCHITECT has a sub-module called RELAY which generates executable Python code that transforms outputs from the optimized schema $S^*$ back to the original schema format $S_{\text{user}}$. The reverse mapping function ensures semantic preservation:
$\text{RELAY}: \mathcal{O}(S^*) \rightarrow \mathcal{O}(S_{user})$,
where $\mathcal{O}(S)$ represents the output space of schema $S$.

RELAY automatically generates and validates this transformation code during the Build Phase, ensuring semantic preservation through:
\begin{enumerate*}[label=(\arabic*)]
\item Automated Python code generation for schema mapping,
\item Sample data pair generation for testing transformations,
\item Iterative refinement until semantic preservation is verified.
\end{enumerate*}
This ensures that downstream systems can continue using original schema formats while benefiting from ARCHITECT's optimizations.

\subsection{\ciemethod:~\ciemethodlong}
SCOPE implements a systematic validation framework with constrained decoding and static rule checking, enabling agents to self-correct through structured reflection. The extraction process with guardrails follows:
$$\hat{y} = \text{Extract}(x, S^*, G_{\text{static}})$$
where $x$ is input text, $S^*$ is the optimized schema, $G_{\text{static}}$ represents static guardrails and validation rules in $S^*$.

\noindent\textbf{Multi-Stage Validation Process:} SCOPE implements a systematic three-stage validation 
that operate as follows:
\begin{enumerate*}[label=(\arabic*)] 
    \item Missing Attribute Check: Verifies that all required fields specified in the schema are present in the extracted output
    \item Grounding Verification: Confirms that extracted values can be found in the original input text
    \item Rule Compliance Check: Validates that extracted values adhere to schema constraints such as patterns, length limits, enumerated values, date format validations, and more.
\end{enumerate*}
Each validation stage returns a status indicator (pass or fail) along with specific error details. When validation fails, SCOPE generates structured reflections that guide the extraction agent toward correct outputs.
\section{Discussion}
\label{sec:discussion}

While PARSE focuses on structured extraction from unstructured text for LLM agent systems, the framework's principles extend to related information extraction domains. Named Entity Recognition (NER)~\cite{li-etal-2020-unified, shrimal-etal-2022-ner}, for instance, can be formulated as schema-based extraction where entity types, spans, and attributes are defined in a structured schema. Multi-agent orchestration frameworks for task automation~\cite{shrimal-etal-2024-marco} face similar challenges in ensuring reliable structured outputs, particularly for tool calling where agents must extract precise parameters from natural language to invoke APIs correctly. In such systems, PARSE's components can enhance execution accuracy: ARCHITECT can optimize tool parameter schemas for clearer LLM comprehension, while SCOPE's reflection-based guardrails can validate parameter extraction and recover from common errors like parameter hallucination or incorrect formatting—challenges explicitly identified in multi-agent systems. This positions PARSE as complementary to agent orchestration frameworks, providing the schema optimization and validation layer needed for reliable tool invocation.

The core insight of PARSE—that schemas themselves can be optimized for LLM consumption rather than treated as static contracts—applies broadly across structured extraction tasks. Whether extracting named entities, tool parameters in agent systems, slot values in dialogue systems, or complex nested structures from web data, the challenge remains: how do we design and refine the structural contracts that guide LLM extraction? PARSE's automated schema optimization through ARCHITECT and systematic validation through SCOPE provide a general framework applicable to these diverse scenarios.

\section{Experiments}
We evaluate PARSE through comprehensive experiments designed to answer three key questions: (1) Does ARCHITECT effectively optimize JSON schemas for LLM consumption? (2) Does SCOPE's reflection-based guardrail system improve extraction reliability? (3) How do these components work together to enable reliable structured information extraction for Software 3.0 applications?

\subsection{Experimental Setup}
\label{sec:experimental_setup}
\noindent\textbf{Datasets:}
We evaluate on three complementary datasets that test different aspects of structured extraction:
\begin{enumerate*}[label=(\arabic*)] 
    \item \textbf{Retail-Conv}: A curated internal dataset of 6 diverse retail conversation schemas with 40 samples each (240 total), designed to test extraction from natural business communications. This dataset captures realistic scenarios where customers describe product issues, requests, and updates in conversational language;
    \item \textbf{Schema-Guided Dialogue (SGD)}~\cite{sgd_data}: A large-scale dataset of 20,000 annotated task-oriented conversations across 20 domains including banking, events, media, and travel.
    \item \textbf{Structured Web Data Extraction (SWDE)}~\cite{swde_data}: A standard benchmark containing 1,600 test samples across 8 verticals (200 test samples per vertical). This dataset tests extraction from semi-structured web content with varied formatting.
\end{enumerate*}

\noindent\textbf{Baseline:} We implement an extraction agent using best prompting practices and constraint decoding with the original user-provided schemas as our baseline. This baseline uses clear instructions, few-shot examples, and standard JSON formatting requests without any guardrails or schema optimization. We evaluate across five LLM variants to ensure robustness: Claude 3.5/3.7 Sonnet, Claude 3.5 Haiku, Llama 4-Maverick, and DeepSeek-R1-671B, spanning diverse architectures and capabilities.

\noindent\textbf{Metrics}: We measure field-level accuracy where all required schema fields must be correctly extracted and properly formatted. This strict metric reflects real-world requirements where partial extraction often provides little value. We also track end-to-end extraction time including all reflection iterations and guardrail checks, providing insight into the practical trade-offs between accuracy and speed. For ARCHITECT evaluation, we analyze schema modification patterns and track accuracy improvements across optimization iterations.

\begin{table*}[htbp]
\centering
\resizebox{0.95\textwidth}{!}{%
\begin{tabular}{@{}lllllllllllll@{}}
\toprule
\multicolumn{1}{c}{Model} &
  \multicolumn{4}{c}{Retail-Conv} &
  \multicolumn{4}{c}{SGD Data} &
  \multicolumn{4}{c}{SWDE Data} \\ \cmidrule(l){2-13} 
 &
  \multicolumn{2}{c}{Original Schema} &
  \multicolumn{2}{c}{ARCHITECT Schema} &
  \multicolumn{2}{c}{Original Schema} &
  \multicolumn{2}{c}{ARCHITECT Schema} &
  \multicolumn{2}{c}{Original Schema} &
  \multicolumn{2}{c}{ARCHITECT Schema} \\
 &
  \multicolumn{1}{c}{\begin{tabular}[c]{@{}c@{}}Baseline\\ Agent\end{tabular}} &
  \multicolumn{1}{c}{SCOPE} &
  \multicolumn{1}{c}{\begin{tabular}[c]{@{}c@{}}Baseline\\ Agent\end{tabular}} &
  \multicolumn{1}{c}{SCOPE} &
  \multicolumn{1}{c}{\begin{tabular}[c]{@{}c@{}}Baseline\\ Agent\end{tabular}} &
  \multicolumn{1}{c}{SCOPE} &
  \multicolumn{1}{c}{\begin{tabular}[c]{@{}c@{}}Baseline\\ Agent\end{tabular}} &
  \multicolumn{1}{c}{SCOPE} &
  \multicolumn{1}{c}{\begin{tabular}[c]{@{}c@{}}Baseline\\ Agent\end{tabular}} &
  \multicolumn{1}{c}{SCOPE} &
  \multicolumn{1}{c}{\begin{tabular}[c]{@{}c@{}}Baseline\\ Agent\end{tabular}} &
  \multicolumn{1}{c}{SCOPE} \\ \midrule
Claude 3.7 Sonnet &
  75 &
  \underline{88.6} &
  77.1 &
  \textbf{90.01} &
  92.3 &
  \underline{94.2} &
  93.1 &
  \textbf{94.8} &
  24.5 &
  \underline{89.2} &
  31.38 &
  \textbf{93.14} \\
Claude 3.5 Sonnet V2 &
  75.5 &
  \underline{88.6} &
  79.8 &
  \textbf{91.7} &
  90.9 &
  \underline{93.9} &
  92.66 &
  \textbf{94.1} &
  25.7 &
  \underline{90.8} &
  33.51 &
  \textbf{92.99} \\
Claude 3.5 Haiku &
  72.5 &
  \underline{85} &
  76.4 &
  \textbf{87.3} &
  88.16 &
  \underline{91.7} &
  89.09 &
  \textbf{93.3} &
  25 &
  \underline{88.3} &
  32.81 &
  \textbf{91.67} \\
Llama 4 Maverick &
  75.5 &
  \underline{84} &
  78.8 &
  \textbf{88.4} &
  83.22 &
  \underline{88.9} &
  84.18 &
  \textbf{91.44} &
  21.3 &
  \underline{85.1} &
  28.99 &
  \textbf{88.32} \\
DeepSeek-R1 &
  76 &
  \underline{91} &
  81.2 &
  \textbf{93.7} &
  87.15 &
  \underline{91.22} &
  89.25 &
  \textbf{92.01} &
  19.3 &
  \underline{82.7} &
  25.13 &
  \textbf{86.16} \\ \bottomrule
\end{tabular}%
}
\caption{Accuracy comparison for Baseline LLMs and SCOPE with and without ARCHITECT schemas on Retail-Conv, SGD and SWDE datasets. \textbf{Bold} indicates best result per model per dataset, \underline{underline} indicates second best.}
\label{tab:master-table}
\end{table*}

\begin{table}[htbp]
\centering
\resizebox{\columnwidth}{!}{%
\begin{tabular}{@{}lcc@{}}
\toprule
                                 & \multicolumn{2}{c}{Retail-Conv}  \\ \cmidrule(l){2-3} 
                                 & ARCHITECT-Claude & ARCHITECT-LLAMA \\ \midrule
\multicolumn{1}{c}{SCOPE-Claude} & \textbf{93.7}             & \underline{91.11}           \\
\multicolumn{1}{c}{SCOPE- LLAMA} & 89.1             & 88.4            \\ \bottomrule
\end{tabular}%
}
\caption{Comparison of performance when Optimising schema using model X in ARCHITECT then applying the schema in SCOPE with model Y. Claude version used is Sonnet 3.5 V2.}
\label{tab:cross-model-perf}
\end{table}

\subsection{Main Results}
\label{subsec:main_results}

\begin{figure}[htbp]
    \centering
    \includegraphics[width=0.48\columnwidth]{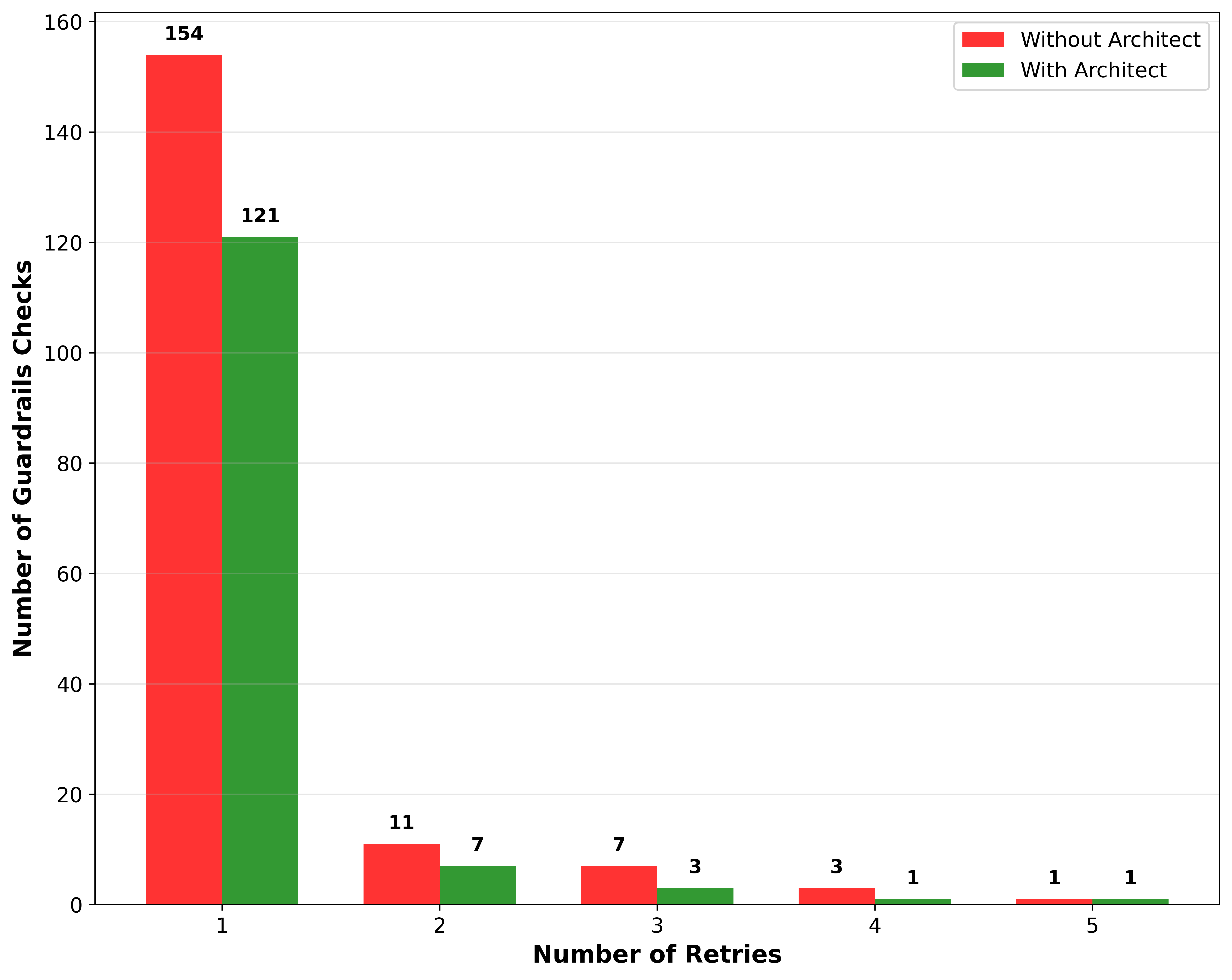}
    \hfill
    \includegraphics[width=0.48\columnwidth]{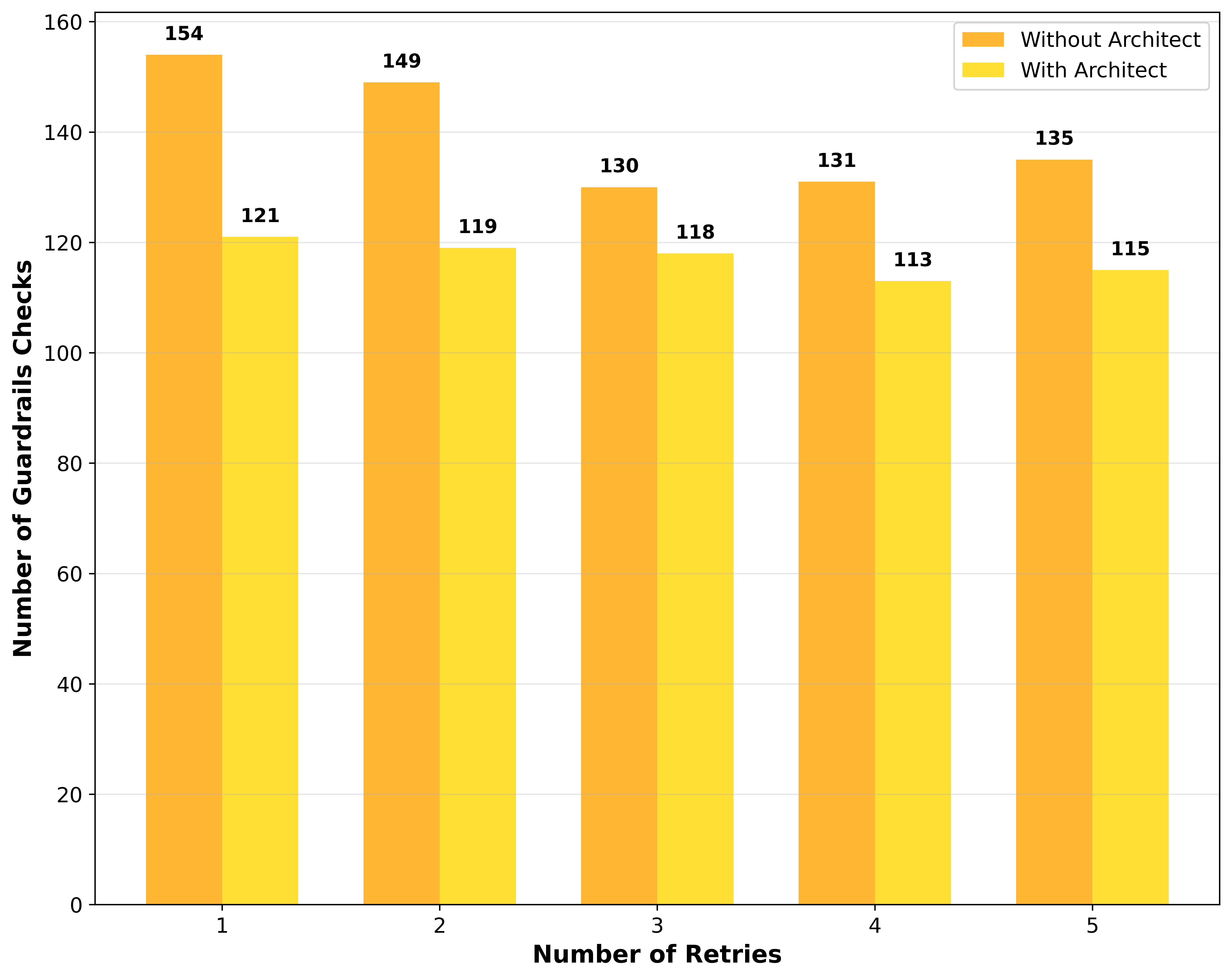}
    \caption{Error recovery with and without reflection and with and without ARCHITECT schemas}
    \label{fig:gr-with-without-ref}
\end{figure}

\begin{figure}[tbh]
  \centering
  \includegraphics[width=0.95\columnwidth]{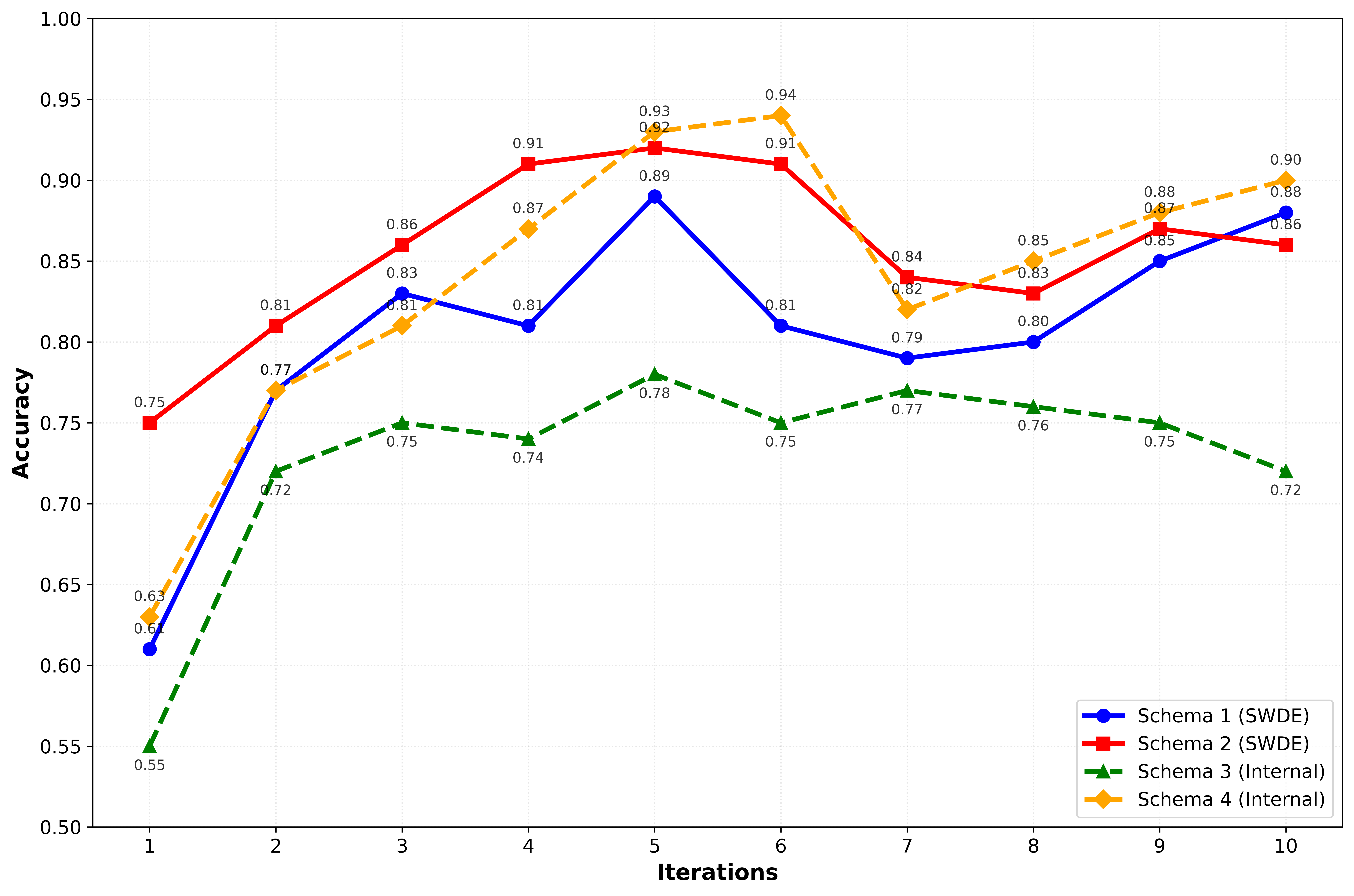}
  \caption{Performance of schemas on validation sets over ARCHITECT optimisation iterations}
  \label{fig:accuracy-architect-iterations}
\end{figure}

\begin{figure}[tbh]
  \centering
  \includegraphics[width=0.95\columnwidth]{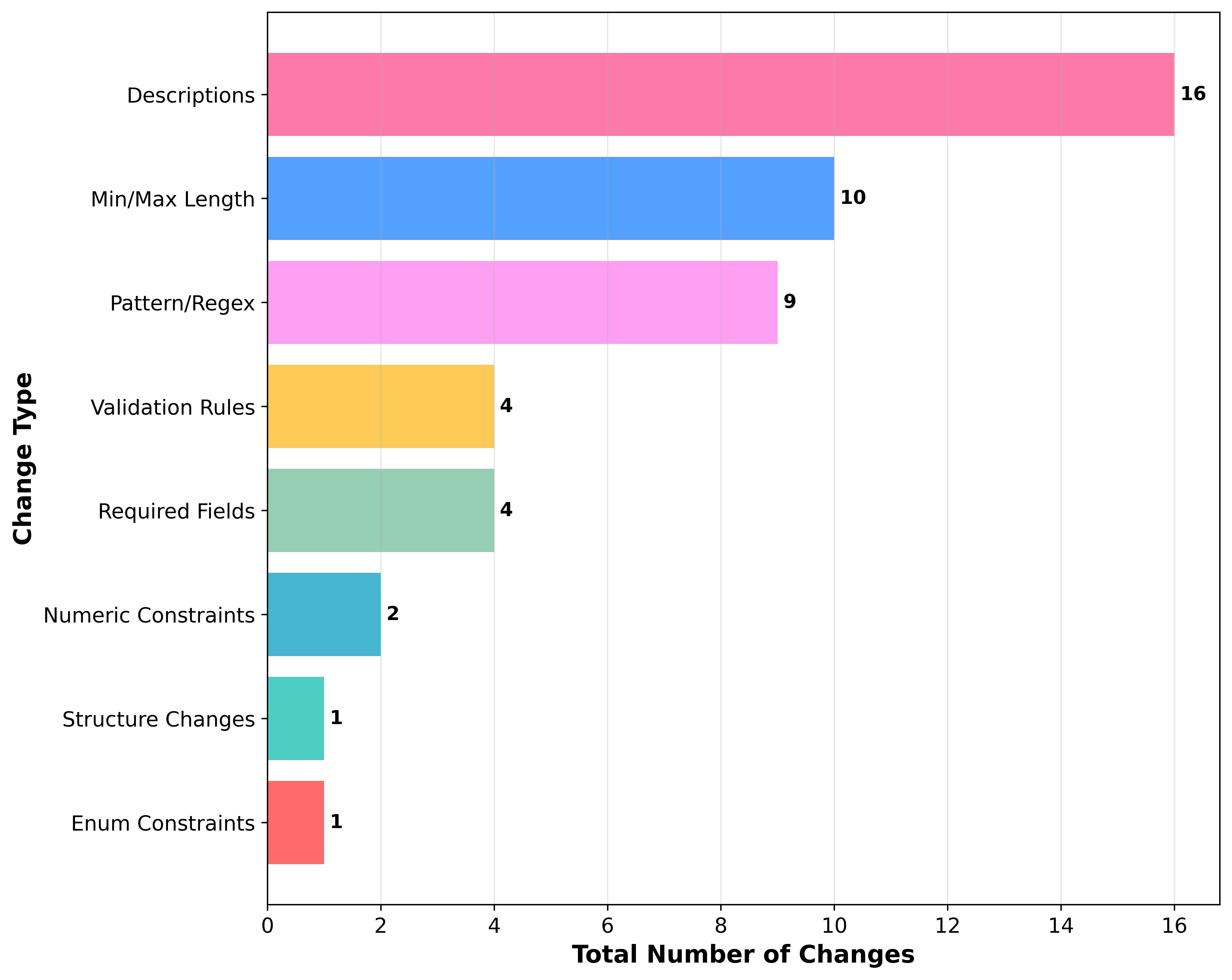}
  \caption{Description of how frequently different types of changes are done through ARCHITECT to optimise a schema}
  \label{fig:change-type-frequency}
\end{figure}

\subsection{Combined System Performance}
\textbf{Performance and Reliability}: Table~\ref{tab:master-table} demonstrates that PARSE achieves substantial improvements across all datasets, with the combined ARCHITECT + SCOPE system showing the strongest performance. The improvements remain consistent across different LLMs suggesting our approach makes sophisticated extraction more accessible across model scales.

\subsection{ARCHITECT Analysis}
\noindent\textbf{Iterative Improvement}: ARCHITECT's schema optimization shows consistent improvement over iterations, with most gains achieved within the first 5-6 iterations before dipping again. This suggests the optimization process efficiently identifies and addresses the most critical schema limitations although doing it for larger durations can lead to overfitting.

\noindent\textbf{Schema Modification Patterns}: Analysis of ARCHITECT's schema changes (Figure~\ref{fig:change-type-frequency}) reveals consistent patterns across datasets with:
\begin{enumerate*}[label=(\arabic*)] 
    \item Entity Description Enhancement (34\%): Adding detailed descriptions and contextual information
    \item Structural Reorganization (55\%): Flattening nested structures and clarifying field relationships  
    \item Validation Rule Addition (0.08\%): Implementing format constraints and enumerated values
    \item Pattern rules additions (3\%): Enforcing strict regex pattern rules to constrain outputs 
\end{enumerate*}

\noindent\textbf{Generalization Across Models}: Table~\ref{tab:cross-model-perf} suggests that schemas optimized using one LLM (e.g., Claude 3.5 Sonnet) maintain their performance benefits when applied with different models (e.g., Llama 4), indicating that ARCHITECT identifies model-agnostic schema improvements rather than model-specific optimizations.

\subsection{SCOPE Guardrail Effectiveness}
\noindent\textbf{Reflection-Based Improvement}: SCOPE's multi-stage validation shows clear benefits over simple retry mechanisms. On average, SCOPE reduces extraction errors by 92\% within the first retry compared to baseline approach that simply re-prompt on failures (Figure~\ref{fig:gr-with-without-ref}). Moreover, ARCHITECT optimised schema's are less prone to have re-tries as shown in the same figure.

\noindent\textbf{Cost-Accuracy Trade-offs}: While SCOPE increases latency by an average of 10.16 due to reflection iterations, the corresponding accuracy gain of +64.7\% on SWDE demonstrates substantial practical value. Importantly, using ARCHITECT-optimized schemas reduces this latency penalty by an average of 4.05s as fewer reflection rounds are needed (Table~\ref{tab:latency}).

\begin{table}[]
\centering
\resizebox{0.8\columnwidth}{!}{%
\begin{tabular}{@{}lrrrr@{}}
\toprule
\multicolumn{1}{c}{Model} & \multicolumn{4}{c}{SWDE Data}                                                             \\ \cmidrule(l){2-5} 
 &
  \multicolumn{2}{c}{\begin{tabular}[c]{@{}c@{}}Latency with \\ Original Schema\end{tabular}} &
  \multicolumn{2}{c}{\begin{tabular}[c]{@{}c@{}}Latency with \\ ARCHITECT Schema\end{tabular}} \\
  \cmidrule(l){2-5} 
 &
  \multicolumn{1}{c}{\begin{tabular}[c]{@{}c@{}}Baseline\\ Agent\end{tabular}} &
  \multicolumn{1}{c}{SCOPE} &
  \multicolumn{1}{c}{\begin{tabular}[c]{@{}c@{}}Baseline\\ Agent\end{tabular}} &
  \multicolumn{1}{c}{SCOPE} \\ \midrule
Claude 3.7 Sonnet         & 9.08                 & 25.58                & 8.80                 & 19.06                \\
Claude 3.5 Sonnet V2      & 9.00                 & 24.46                & 8.83                 & 19.54                \\
Claude 3.5 Haiku          & 8.99                 & 22.26                & 8.68                 & 14.99                \\
Llama 4 Maverick          & 0.98                 & 2.30                 & 0.93                 & 1.97                 \\
DeepSeek-R1               & 3.01                 & 11.99                & 2.91                 & 10.01                \\ \midrule
                          & \multicolumn{1}{c}{} & \multicolumn{1}{c}{} & \multicolumn{1}{c}{} & \multicolumn{1}{l}{}
\end{tabular}%
}
\caption{Impact on latency for extraction with using SCOPE and ARCHITECT}
\label{tab:latency}
\end{table}

\subsection{Qualitative Analysis}
ARCHITECT consistently transforms ambiguous field descriptions into precise, context-rich specifications. For example, a generic "price" field becomes "newPriceValue: The specific numerical price value for the product update, excluding currency symbols" with additional validation rules for format and range constraints.

SCOPE's reflection mechanism effectively identifies and corrects common extraction errors. In cases where the baseline agent confused similar entities (e.g., "old price" vs. "new price"), SCOPE's grounding verification caught the error and guided correction through structured reflection.

\noindent\textbf{SWDE Dataset Complexity:} The substantial improvements on SWDE (up to 64.7\%) stem from the dataset's HTML structure requiring precise extraction with specific formatting. Table~\ref{tab:swde-example} illustrates a representative case where the baseline struggles with HTML noise while PARSE's optimized schema and guardrails enable accurate extraction. The baseline schema's generic \texttt{"model": \{"type": "string"\}} provides insufficient guidance for parsing complex HTML, while ARCHITECT's optimized schema adds detailed descriptions and pattern constraints that help the LLM focus on relevant content. SCOPE's grounding verification then ensures extracted values match the source text, preventing hallucinations common in HTML extraction tasks.

Refer to Appendix for some qualitative samples on schema optimisations done by ARCHITECT.

\begin{table}[t]
\centering
\resizebox{\columnwidth}{!}{%
\begin{tabular}{@{}p{3cm}p{5.7cm}@{}}
\toprule
\textbf{Component} & \textbf{Content} \\ 
\midrule
\textbf{Input HTML} & \texttt{<title>2010 Subaru Legacy 2.5 i 4dr Sedan</title>} (truncated) \\
\midrule
\textbf{Baseline Schema} & \texttt{\{"model": \{"type": "string"\}\}} \\
\midrule
\textbf{PARSE Schema} & \texttt{\{"model": \{"description": "Full model name including year, make and model", "pattern": "\^{}(19[5-9][0-9]|20[0-2][0-9]) [A-Za-z0-9 -+]+\$"...\}\}} \\
\midrule
\textbf{Expected Result} & "2010 Subaru Legacy" \\
\midrule
\textbf{Baseline Result} & "2010 Subaru Legacy 2.5 i 4dr Sedan" \\
\midrule
\textbf{PARSE Result} & "2010 Subaru Legacy" \\
\bottomrule
\end{tabular}%
}
\caption{SWDE extraction example demonstrating why PARSE achieves substantial improvements on HTML-structured data. The baseline extracts excessive detail due to insufficient schema guidance, while PARSE's optimized schema with pattern constraints and SCOPE's validation ensure precise extraction.}
\label{tab:swde-example}
\end{table}
\section*{Conclusion}
We introduced PARSE, a comprehensive framework that addresses structured information extraction challenges through schema optimization and reflection-based guardrails. ARCHITECT automatically refines JSON schemas for LLM consumption while SCOPE ensures reliable extraction through multi-stage validation. Our evaluation across three datasets demonstrates substantial improvements, with up to 64.7\% accuracy gains on SWDE and 92\% error reduction within first retry, establishing PARSE as an effective solution for reliable LLM agent systems in Software 3.0 applications.

\section*{Limitations}
Our approach has several important limitations that future work should address. The iterative refinement process in ARCHITECT can be computationally expensive for complex schemas with many attributes. Each refinement iteration requires synthetic data generation, extraction evaluation, and failure analysis, creating potential scalability bottlenecks for large-scale deployments.

ARCHITECT's optimization quality depends heavily on the availability and representativeness of seed datasets. For entirely new domains or rapidly evolving schema requirements, obtaining sufficient high-quality seed data can be challenging. Furthermore, our approach assumes relatively static schema structures that can be optimized offline, which can be challenging where schemas evolve continuously.

Several promising directions emerge from this work. Multi-modal extension represents a natural evolution, where schemas could be optimized for extraction from both textual and visual content. The same principle of treating schemas as optimizable interfaces applies to multi-modal scenarios, but would require extending the validation framework to handle cross-modal grounding.

\bibliography{references}

\appendix
\section{Appendix}

\subsection{Qualitative Samples of SCOPE predictions}

Table \ref{tab:qualitative-data} gives an overview of how SCOPE performs against Baselines. Notice that SCOPE is able handle longer contexts better while 

\definecolor{mygreen}{RGB}{144,238,144} 
\definecolor{myred}{RGB}{255,160,122}   
\definecolor{myblue}{RGB}{173,216,230}  
\renewcommand{\arraystretch}{1.3}

\begin{table*}[t]
\centering
\resizebox{\textwidth}{!}{%
\begin{tabular}{@{}p{1.8cm}p{2.2cm}p{8.2cm}p{4cm}p{4.5cm}@{}}
\toprule
\textbf{Dataset} & \textbf{Input Schema} & \textbf{Context} & \textbf{SCOPE} & \textbf{Baseline} \\ 
\midrule

\textbf{SGD} &
Restaurant\_2 &
\textbf{[USER]}: I want to have a table booked for me at a restaurant, because I want to have a meal. \newline
\textbf{[ASSISTANT]}: Can you tell me the restaurant you are interested to have a table booked at? \newline
\textbf{[USER]}: Can you check if you can make a booking of a table at Castle Rock? \newline
\textbf{[ASSISTANT]}: Can you tell me the time that you want to have the table booked for you? \newline
\textbf{[USER]}: I prefer to have the booking be made for \colorbox{myblue}{12:00}. \newline
\textbf{[ASSISTANT]}: Is there a specific location you are looking for? \newline
\textbf{[USER]}: I am interested in restaurants which are located in \colorbox{mygreen}{Livermore}. \newline
\textbf{[ASSISTANT]}: Can you please confirm that you want to book a table for 2 people at Castle Rock Restaurant in Livermore, and that you want the table to be booked for you for 12 pm on March 1st? \newline
\textbf{[USER]}: That is not right. Please make a booking at \colorbox{myred}{Mcdonald's}. \newline
\textellipsis &
\{ \newline
\hspace{0.5cm}"ReserveRestaurant": {[} \newline
\hspace{1cm}\{"location": \colorbox{mygreen}{"Livermore"}\}, \newline
\hspace{1cm}\{"restaurant\_name": \colorbox{myred}{"Mcdonald's"}\}, \newline
\hspace{1cm}\{"time": \colorbox{myblue}{"12 pm"}\} \newline
\hspace{0.5cm}{]} \newline
\} &
\{ \newline
\hspace{0.5cm}"ReserveRestaurant": {[} \newline
\hspace{1cm}\{ \newline
\hspace{1.5cm}"restaurant\_name": null, \newline
\hspace{1.5cm}"location": \colorbox{mygreen}{"Livermore"}, \newline
\hspace{1.5cm}"time": \colorbox{myblue}{"12:00"}, \newline
\hspace{1.5cm}"number\_of\_seats": null, \newline
\hspace{1.5cm}"date": null \newline
\hspace{1cm}\} \newline
\hspace{0.5cm}{]} \newline
\} \\
\midrule
\textbf{SGD} &
Payment\_1, Trains\_1, Weather\_1 &
\textbf{[USER]}: I am looking to make a payment. \newline
\textbf{[ASSISTANT]}: What is the name of the contact and from what account? \newline
\textbf{[USER]}: I'd like it from my checking account and the name is Mahmoud. \newline
\textbf{[ASSISTANT]}: How much will it be? \newline
\textbf{[USER]}: The amount is \$180. \newline
\textbf{[ASSISTANT]}: Please confirm you're making a transaction for \$180 from your debit card with Mahmoud and it will be visible to the public. \newline
\textbf{[USER]}: That is correct. \newline
\textbf{[ASSISTANT]}: Your payment has been confirmed and the receiver should see the money soon. \newline
\textbf{[USER]}: Can you help me find a train for the \colorbox{myblue}{12th of March}? \newline
\textbf{[ASSISTANT]}: From which city are you leaving from? \newline
\textbf{[USER]}: I'm planning to leave from \colorbox{mygreen}{Sacramento, CA}. \newline
\textbf{[ASSISTANT]}: Where are you going? \newline
\textbf{[USER]}: I'm trying to get to \colorbox{myred}{San Fran}. \newline
\textbf{[ASSISTANT]}: I found a train leaving at \colorbox{myblue}{7:30 am} for \colorbox{myblue}{\$47}. \newline
\textbf{[USER]}: Which station does this train leave from? \newline
\textbf{[ASSISTANT]}: It is leaving from \colorbox{mygreen}{Sacramento Valley Station}. \newline
\textbf{[USER]}: That works. How will the weather be in \colorbox{myred}{San Fran} on the \colorbox{myblue}{14th}? \newline
\textellipsis &
\{ \newline
\hspace{0.5cm}"GetWeather": {[} \{"city": \colorbox{myred}{"San Fran"}, "date": \colorbox{myblue}{"the 14th"}\} {]}, \newline
\hspace{0.5cm}"FindTrains": {[} \newline
\hspace{1cm}\{"date\_of\_journey": \colorbox{myblue}{"12th of March"}\}, \newline
\hspace{1cm}\{"from": \colorbox{mygreen}{"Sacramento, CA"}\}, \newline
\hspace{1cm}\{"journey\_start\_time": \colorbox{myblue}{"7:30 am"}\}, \newline
\hspace{1cm}\{"to": \colorbox{myred}{"San Fran"}\} \newline
\hspace{0.5cm}{]} \newline
\} &
\{ \newline
\hspace{0.5cm}"GetWeather": {[} \{"city": \colorbox{myred}{"San Francisco"}, "date": \colorbox{myblue}{"March 14th"}\} {]} \newline
\} \\
\midrule
\textbf{Retail-Conv} &
Sales Data &
\textbf{[USER]}: Get sales data for \colorbox{myred}{"PRODUCT123"} in \colorbox{mygreen}{Italy} \newline
\textbf{[ASSISTANT]}: I need a specific timeframe to fetch the sales data. Could you please specify when you want to see the sales for? For example: last week, MTD, or a specific date range? & 
\{NULL\} &
\{"sales": {[} \{"asin": \colorbox{myred}{"PRODUCT123"}, "timeframe": \colorbox{myblue}{"MTD"}, "marketplace": \colorbox{mygreen}{"IT"}\} {]} \newline
\} \\
\midrule
\textbf{Retail-Conv} &
Price update &
\textbf{[USER]}: For \colorbox{myred}{PRODUCT234}, the old price is \colorbox{myblue}{29.99 USD} and I want to maybe change it to either \colorbox{myblue}{39.99} or \colorbox{myblue}{49.99 USD} &
\{NULL\} &
\{"list\_of\_asin\_price\_vendor": {[}\{ \newline
\hspace{1.5cm}"asin": \colorbox{myred}{"PRODUCT234"}, \newline
\hspace{1.5cm}"new\_list\_price\_value": \colorbox{myblue}{49.99}, \newline
\hspace{1.5cm}"new\_list\_price\_currency": "USD"\}{]}\} \\
\bottomrule
\end{tabular}%
}
\caption{Qualitative comparison between SCOPE and baselines predictions (both with ARCHITECT) on Claude 3.5 Sonnet V2.}
\label{tab:qualitative-data}
\end{table*}

\subsection{Qualitative Samples of ARCHITECT's optimisation}
Table~\ref{tab:qualitative-architect-schemas} shows the qualitative example of how ARCHITECT's optimisation process updates the JSON schema across 5 iterations for various attributes in SWDE dataset.

\definecolor{mygreen}{RGB}{200,230,200}  
\definecolor{myblue}{RGB}{200,220,255}   
\definecolor{myyellow}{RGB}{255,245,200} 

\renewcommand{\arraystretch}{1.3}

\begin{table*}[t]
\centering
\small
\resizebox{\textwidth}{!}{%
\begin{tabular}{@{}p{1cm}p{4cm}p{4cm}p{4cm}p{4cm}@{}}
\toprule
\textbf{Iteration} & \textbf{automobile\_model\_year} & \textbf{price} & \textbf{engine\_type} & \textbf{fuel\_economy} \\ 
\midrule

1 &
\begin{tabular}[t]{@{}p{3.8cm}@{}}
\texttt{\{} \\
\hspace{0cm}\texttt{"type": "array",} \\
\hspace{0cm}\texttt{"description": "List of} \\
\hspace{0cm}\texttt{automobile models with} \\
\hspace{0cm}\texttt{their corresponding years",} \\
\hspace{0cm}\texttt{"items": \{} \\
\hspace{0.2cm}\texttt{"type": "object",} \\
\hspace{0.2cm}\texttt{"properties": \{} \\
\hspace{0.4cm}\texttt{"model": \{} \\
\hspace{0.6cm}\texttt{"type": "string"} \\
\hspace{0.4cm}\texttt{\}} \\
\texttt{\}}
\texttt{\}}
\texttt{\}}
\end{tabular} &
\begin{tabular}[t]{@{}p{3.8cm}@{}}
\texttt{\{} \\
\hspace{0cm}\texttt{"type": "array",} \\
\hspace{0cm}\texttt{"description": "List of} \\
\hspace{0cm}\texttt{price information",} \\
\hspace{0cm}\texttt{"items": \{} \\
\hspace{0.2cm}\texttt{"type": "object",} \\
\hspace{0.2cm}\texttt{"properties": \{} \\
\hspace{0.4cm}\texttt{"price": \{} \\
\hspace{0.6cm}\texttt{"type": "string",} \\
\hspace{0.6cm}\texttt{"description":} \\
\hspace{0.6cm}\texttt{"Currency symbol +} \\
\hspace{0.6cm}\texttt{numeric value,} \\
\hspace{0.6cm}\texttt{eg: \$19,995"} \\
\texttt{\}}
\texttt{\}}
\texttt{\}}
\texttt{\}}
\end{tabular} &
\begin{tabular}[t]{@{}p{3.8cm}@{}}
\texttt{\{} \\
\hspace{0cm}\texttt{"type": "array",} \\
\hspace{0cm}\texttt{"description": "List of} \\
\hspace{0cm}\texttt{engine types",} \\
\hspace{0cm}\texttt{"items": \{} \\
\hspace{0.2cm}\texttt{"type": "object",} \\
\hspace{0.2cm}\texttt{"properties": \{} \\
\hspace{0.4cm}\texttt{"engine": \{} \\
\hspace{0.6cm}\texttt{"type": "string"} \\
\texttt{\}}
\texttt{\}}
\texttt{\}}
\texttt{\}}
\end{tabular} &
\begin{tabular}[t]{@{}p{3.8cm}@{}}
\texttt{\{} \\
\hspace{0cm}\texttt{"type": "array",} \\
\hspace{0cm}\texttt{"items": \{} \\
\hspace{0.2cm}\texttt{"type": "object",} \\
\hspace{0.2cm}\texttt{"properties": \{} \\
\hspace{0.4cm}\texttt{"fuel": \{} \\
\hspace{0.6cm}\texttt{"type": "string"} \\
\texttt{\}}
\texttt{\}}
\texttt{\}}
\texttt{\}}
\end{tabular} \\

\midrule

3 &
\begin{tabular}[t]{@{}p{3.8cm}@{}}
\texttt{\{} \\
\hspace{0cm}\texttt{"type": "array",} \\
\hspace{0cm}\texttt{"description":} \\
\hspace{0cm}\colorbox{myyellow}{\texttt{"List of automobile}} \\
\hspace{0cm}\colorbox{myyellow}{\texttt{models with full year,}} \\
\hspace{0cm}\colorbox{myyellow}{\texttt{make and model"}}\texttt{,} \\
\hspace{0cm}\texttt{"items": \{} \\
\hspace{0.2cm}\texttt{"type": "object",} \\
\hspace{0.2cm}\texttt{"required":} \\
\hspace{0.2cm}\colorbox{mygreen}{\texttt{["model"]}}\texttt{,} \\
\hspace{0.2cm}\texttt{"properties": \{} \\
\hspace{0.4cm}\texttt{"model": \{} \\
\hspace{0.6cm}\texttt{"type": "string",} \\
\hspace{0.6cm}\texttt{"pattern":} \\
\hspace{0.6cm}\colorbox{myblue}{\texttt{"\textasciicircum[0-9]\{4\}}} \\
\hspace{0.6cm}\colorbox{myblue}{\texttt{[A-Za-z0-9 -]+"}}\texttt{,} \\
\hspace{0.6cm}\texttt{"minLength": 10} \\
\texttt{\}}
\texttt{\}}
\texttt{\}}
\texttt{\}}
\end{tabular} &
\begin{tabular}[t]{@{}p{3.8cm}@{}}
\texttt{\{} \\
\hspace{0cm}\texttt{"type": "array",} \\
\hspace{0cm}\texttt{"description": "List of} \\
\hspace{0cm}\texttt{price information",} \\
\hspace{0cm}\texttt{"items": \{} \\
\hspace{0.2cm}\texttt{"type": "object",} \\
\hspace{0.2cm}\texttt{"required":} \\
\hspace{0.2cm}\colorbox{mygreen}{\texttt{["price"]}}\texttt{,} \\
\hspace{0.2cm}\texttt{"properties": \{} \\
\hspace{0.4cm}\texttt{"price": \{} \\
\hspace{0.6cm}\texttt{"type": "string",} \\
\hspace{0.6cm}\texttt{"pattern":} \\
\hspace{0.6cm}\colorbox{myblue}{\texttt{"\textasciicircum[\$\texteuro\textyen]}} \\
\hspace{0.6cm}\colorbox{myblue}{\texttt{[0-9,.]+}} \\
\hspace{0.6cm}\colorbox{myblue}{\texttt{(?:k|K)?\$"}}\texttt{,} \\
\hspace{0.6cm}\texttt{"minLength": 2} \\
\texttt{\}}
\texttt{\}}
\texttt{\}}
\texttt{\}}
\end{tabular} &
\begin{tabular}[t]{@{}p{3.8cm}@{}}
\texttt{\{} \\
\hspace{0cm}\texttt{"type": "array",} \\
\hspace{0cm}\texttt{"description": "List of} \\
\hspace{0cm}\texttt{engine/powertrain} \\
\hspace{0cm}\texttt{configurations",} \\
\hspace{0cm}\texttt{"items": \{} \\
\hspace{0.2cm}\texttt{"type": "object",} \\
\hspace{0.2cm}\texttt{"required":} \\
\hspace{0.2cm}\colorbox{mygreen}{\texttt{["engine"]}}\texttt{,} \\
\hspace{0.2cm}\texttt{"properties": \{} \\
\hspace{0.4cm}\texttt{"engine": \{} \\
\hspace{0.6cm}\texttt{"type": "string",} \\
\hspace{0.6cm}\texttt{"pattern":} \\
\hspace{0.6cm}\colorbox{myblue}{\texttt{"\textasciicircum[0-9.]+L}} \\
\hspace{0.6cm}\colorbox{myblue}{\texttt{(?:$\backslash$s+[A-Za-z0-9-}} \\
\hspace{0.6cm}\colorbox{myblue}{\texttt{$\backslash$s]+)*\$"}}\texttt{,} \\
\hspace{0.6cm}\texttt{"minLength": 3} \\
\texttt{\}}
\texttt{\}}
\texttt{\}}
\texttt{\}}
\end{tabular} &
\begin{tabular}[t]{@{}p{3.8cm}@{}}
\texttt{\{} \\
\hspace{0cm}\texttt{"type": "array",} \\
\hspace{0cm}\texttt{"description": "List of} \\
\hspace{0cm}\texttt{fuel economy ratings",} \\
\hspace{0cm}\texttt{"items": \{} \\
\hspace{0.2cm}\texttt{"type": "object",} \\
\hspace{0.2cm}\texttt{"required":} \\
\hspace{0.2cm}\colorbox{mygreen}{\texttt{["fuel"]}}\texttt{,} \\
\hspace{0.2cm}\texttt{"properties": \{} \\
\hspace{0.4cm}\texttt{"fuel": \{} \\
\hspace{0.6cm}\texttt{"type": "string",} \\
\hspace{0.6cm}\texttt{"minLength": 5,} \\
\hspace{0.6cm}\texttt{"condition": "Must} \\
\hspace{0.6cm}\texttt{include numeric value} \\
\hspace{0.6cm}\texttt{with unit (MPG, MPGe)"} \\
\texttt{\}}
\texttt{\}}
\texttt{\}}
\texttt{\}}
\end{tabular} \\

\midrule

5 &
\begin{tabular}[t]{@{}p{3.8cm}@{}}
\texttt{\{} \\
\hspace{0cm}\texttt{"type": "array",} \\
\hspace{0cm}\texttt{"description":} \\
\hspace{0cm}\colorbox{myyellow}{\texttt{"Full model name}} \\
\hspace{0cm}\colorbox{myyellow}{\texttt{including year, make,}} \\
\hspace{0cm}\colorbox{myyellow}{\texttt{model, hyphens and}} \\
\hspace{0cm}\colorbox{myyellow}{\texttt{plus signs"}}\texttt{,} \\
\hspace{0cm}\texttt{"items": \{} \\
\hspace{0.2cm}\texttt{"type": "object",} \\
\hspace{0.2cm}\texttt{"required":} \\
\hspace{0.2cm}\colorbox{mygreen}{\texttt{["model"]}}\texttt{,} \\
\hspace{0.2cm}\texttt{"properties": \{} \\
\hspace{0.4cm}\texttt{"model": \{} \\
\hspace{0.6cm}\texttt{"type": "string",} \\
\hspace{0.6cm}\texttt{"pattern":} \\
\hspace{0.6cm}\colorbox{myblue}{\texttt{"\textasciicircum[0-9.]+L}} \\
\hspace{0.6cm}\colorbox{myblue}{\texttt{(?:$\backslash$s+[A-Za-z0-9-}} \\
\hspace{0.6cm}\colorbox{myblue}{\texttt{$\backslash$s$\backslash$+]+)*\$"}}\texttt{,} \\
\hspace{0.6cm}\texttt{"minLength": 10} \\
\texttt{\}}
\texttt{\}}
\texttt{\}}
\texttt{\}}
\end{tabular} &
\begin{tabular}[t]{@{}p{3.8cm}@{}}
\texttt{\{} \\
\hspace{0cm}\texttt{"type": "array",} \\
\hspace{0cm}\texttt{"required":} \\
\hspace{0cm}\colorbox{mygreen}{\texttt{["price"]}}\texttt{,} \\
\hspace{0cm}\texttt{"items": \{} \\
\hspace{0.2cm}\texttt{"price": \{} \\
\hspace{0.4cm}\texttt{"type": "string",} \\
\hspace{0.4cm}\texttt{"pattern":} \\
\hspace{0.4cm}\colorbox{myblue}{\texttt{"\textasciicircum[\$\texteuro\textyen\pounds]}} \\
\hspace{0.4cm}\colorbox{myblue}{\texttt{[0-9,.]+}} \\
\hspace{0.4cm}\colorbox{myblue}{\texttt{(?:k|K)?\$"}}\texttt{,} \\
\hspace{0.4cm}\texttt{"minLength": 2} \\
\texttt{\}}
\texttt{\}}
\texttt{\}}
\end{tabular} &
\begin{tabular}[t]{@{}p{3.8cm}@{}}
\texttt{\{} \\
\hspace{0cm}\texttt{"type": "array",} \\
\hspace{0cm}\texttt{"required":} \\
\hspace{0cm}\colorbox{mygreen}{\texttt{["engine"]}}\texttt{,} \\
\hspace{0cm}\texttt{"items": \{} \\
\hspace{0.2cm}\texttt{"engine": \{} \\
\hspace{0.4cm}\texttt{"type": "string",} \\
\hspace{0.4cm}\texttt{"pattern":} \\
\hspace{0.4cm}\colorbox{myblue}{\texttt{"\textasciicircum[0-9.]+L}} \\
\hspace{0.4cm}\colorbox{myblue}{\texttt{(?:$\backslash$s+[A-Za-z0-9-}} \\
\hspace{0.4cm}\colorbox{myblue}{\texttt{$\backslash$s]+)*\$"}}\texttt{,} \\
\hspace{0.4cm}\texttt{"minLength": 3} \\
\hspace{0.2cm}\texttt{\}} \\
\hspace{0cm}\texttt{\}} \\
\texttt{\}}
\end{tabular} &
\begin{tabular}[t]{@{}p{3.8cm}@{}}
\texttt{\{} \\
\hspace{0cm}\texttt{"type": "array",} \\
\hspace{0cm}\texttt{"required":} \\
\hspace{0cm}\colorbox{mygreen}{\texttt{["fuel"]}}\texttt{,} \\
\hspace{0cm}\texttt{"items": \{} \\
\hspace{0.2cm}\texttt{"fuel": \{} \\
\hspace{0.4cm}\texttt{"type": "string",} \\
\hspace{0.4cm}\texttt{"minLength": 5,} \\
\hspace{0.4cm}\texttt{"condition": "Must} \\
\hspace{0.4cm}\texttt{include numeric value} \\
\hspace{0.4cm}\texttt{with unit and flexible} \\
\hspace{0.4cm}\texttt{separators"} \\
\hspace{0.2cm}\texttt{\}} \\
\hspace{0cm}\texttt{\}} \\
\texttt{\}}
\end{tabular} \\

\bottomrule
\end{tabular}%
}
\caption{Evolution of SWDE Auto schema over iterations. \colorbox{mygreen}{Green} highlights new required fields, \colorbox{myblue}{Blue} highlights pattern changes, \colorbox{myyellow}{Yellow} highlights description updates.}
\label{tab:qualitative-architect-schemas}
\end{table*}

\subsection{RELAY: Ensuring Backward Compatibility}
RELAY (Reverse Engineering Layer for Automated Yoking) maintains compatibility between ARCHITECT-optimized schemas and original user schemas through automated code generation. When ARCHITECT transforms a schema, RELAY simultaneously creates Python functions that map outputs between formats.

\textbf{Example Transformation:}
\begin{itemize}
\item Original schema: \texttt{\{"price": \{"type": "string"\}\}}
\item Optimized schema: \texttt{\{"currency\_symbol": \{"type": "string", "pattern": "[\$€\yen\pounds]"\}, "price\_value": \{"type": "number", "minimum": 0\}\}}
\item Generated RELAY code:
\begin{lstlisting}[language=Python]
def reverse_transform(optimized_data):
    return {
        "price": f"{optimized_data['currency_symbol']}"
                 f"{optimized_data['price_value']}"
    }
\end{lstlisting}
\end{itemize}

\textbf{Validation Process:} RELAY generates sample data pairs and tests transformation code iteratively. If semantic preservation errors are detected, it refines the transformation logic until accuracy is achieved. Production deployments show consistent transformation accuracy without conversion errors.

\newtcolorbox{promptboxfull}[1][]{
  enhanced,
  colback=gray!5,
  colframe=gray!30!black,
  title=#1,
  arc=1mm,
  boxrule=0.3pt,
  left=2mm,
  right=2mm,
  boxsep=1mm,
  width=\textwidth,
  breakable
}
\onecolumn
\subsection{ARCHITECT Agent Prompt}
\begin{promptboxfull}[Schema Generator Agent Prompt]
\small
\begin{verbatim}
You are a specialized schema generation agent that creates precise schemas for information extraction. 

The contract must be a valid json schema.

The contracts must strictly adhere to a json format.

Task description: \$task
Use the task description to understand:
- What variables should be included in the schema
- Their data types and structures
- Any constraints or patterns they follow

Please generate the following:
1. A json schema that is optimal for extracting values of attributes mentioned in the schema.
A schema is considered optimal for extraction if it necessarily fulfills the following conditions:

1. The generated schema must be as concise as possible. This is important to 
ensure least latency for downstream extraction.
2. The schema must contain non-conflicting attributes which are non-ambiguous 
and do not cause any confusion while performing extraction. 
This can be achieved through:
a) Coming up with dis-similar names to prevent any confusion
b) Creating clear descriptions for the attribute
c) Creating conditions, rules wherever necessary

Requirements:
- Follow the json format strictly. Keep it as simple as possible.
- Include clear descriptions for each attribute
- Add proper constraints (e.g., required fields, patterns, length, enums) 
where appropriate
- Consider edge cases and error scenarios
- Use appropriate data types optimized for the use case

Do NOT add any extra attributes apart from the ones mentioned in the task description
First, analyze the task thoroughly in <thinking></thinking> tags, considering:
- What attributes are needed to complete the task?
- What constraints or validations apply?
- What types best represent each piece of data?
- Are there any optional parameters to consider?
Return your generated schema in <json_schema></json_schema> tags.
\end{verbatim}
\end{promptboxfull}
\twocolumn

\begin{promptboxfull}[Synthetic Test Data Generator Prompt]
\small
\begin{verbatim}
You are an expert in creating challenging datasets that expose flaws in attribute extraction systems. 
Your task is to generate diverse, edge-case rich examples that will thoroughly test 
and potentially break a JSON schema-based attribute extraction system.
User provided json schema: \$schema
The user defined the attribute extraction task as: \$task
## Instructions
1. I will provide you with a JSON schema that defines attributes to be extracted.
2. Analyze this schema carefully to identify potential weaknesses, edge cases, and ambiguities.
3. Generate a comprehensive dataset of examples designed to challenge the extraction system.
4. For each example, provide:
   - Input text containing the information to be extracted
   - Expected output (what the correct extraction should be, You can give an 
   empty output as well whenever required)
      - If all required fields in the schema are not present, the ground truth 
      should be empty
   - Description of why this example is challenging
5. Consider creating multiple message conversations with back and forth 
between USER and ASSISTANT to make the dataset more complex.
Think hard before generating your samples. Include your thinking 
in <thinking></thinking> tags.
## Guidelines for Creating Adversarial Examples
Create examples that target these vulnerabilities:
1. **Contextual ambiguity**:
   - Multiple potential matches for the same attribute
   - Contradictory information
   - Attribute values embedded in complex sentences
2. **Structural challenges**:
   - Nested information
   - Lists containing relevant attributes
   - Tabular data represented in text
3. **Semantic traps**:
   - Similar but incorrect values
   - Information that appears to match the schema but doesn't
   - Deliberate misinformation or red herrings
4. **Linguistic complexity**:
   - Jargon and domain-specific terminology
   - Colloquial expressions of values
   - Indirect references
5. **Error conditions**:
   - Malformed inputs
   - Missing required attributes
   - Data type mismatches
Create samples where the current schema is not sufficient to handle the input, 
then give ground truth as "INSUFFICIENT_SCHEMA".
## Output Format
For each generated example, structure your output as follows:
<example>
### Example [number]
<input_text>
</input_text>
<ground_truth>
{
  "attribute1": "value1",
  ...
}
</ground_truth>

<challenge> [boundary case/ambiguity/etc.] </challenge>
</example>
Generate atleast TEN diverse and challenging examples based on the provided 
schema, each targeting different vulnerabilities.

You can use the following user provided samples: user_samples
\end{verbatim}
\end{promptboxfull}
\twocolumn

\begin{promptboxfull}[Schema refinement Prompt]
\small
\begin{verbatim}
You are a schema refinement agent specializing in improving JSON schemas for attribute extraction tasks. 
Your goal is to analyze evaluation results and modify the schema to prevent failure
cases while maintaining accuracy on successful extractions.
1. Task which the json schema should be based on: $task
2. The original JSON schema: $schema
3. A set of evaluation samples containing: $eval_samples
For your context, the schema will be passed to the following tool to perform 
extraction:
- The tool uses a json schema and passes it to a LLM agent
- This schema defines attributes and their inter-dependent relationships
- The schema contains information about each attribute that has to be extracted 
from an unstructured context.
- The LLM agent fills the schema with the relevant values from the context
- The agent also has validation guardrails (applied in the same order): 
1. missing attribute check - which checks if the required attributes in the 
json schema are missing from the user provided input context or not.
2. Grounding - which checks if the values predicted are present in the 
context or not.
3. rules mismatch - which checks if the attribute value adheres to the 
specified rules: which can be minLength, maxLength, enums, regex patterns
Analyze the success and failure patterns to propose a refined schema that 
addresses the issues identified.
You can change the entire structure of the schema if you think there is a 
more optimal approach as well. 
The contracts must strictly adhere to a json format. You can only use the following 
fields for an attirbute in the schema: 
[name, description, type, enum, properties, title, pattern, 
minLength, maxLength, condition]. 
You CANNOT USE these keys: if, else, anyof, allof.
Grounding of dates happen through LLM based prompting to handle different 
prompts if each date attribute has the following fiels in it:
"date": {
   "allowed_date_formats": [
      "MM/DD/YYYY"
   ],
   "delimiter": "/"
}
"condition" key in the schema should contain a natural language condition only.
Do NOT add any extra attributes apart from the ones mentioned in the 
task description
## Output Format
Present your analysis and solution as follows:
<failure_analysis>
[Detailed analysis of failure patterns observed in the evaluation samples]
</failure_analysis>
<issues>
[List of specific schema issues that need to be addressed]
</issues>
<improvements>
[Specific changes recommended to address each issue]
</improvements>
<refined_schema>
{
  // Your complete refined schema here
}
<refined_schema>
Ensure your refined schema is backward compatible with successful 
cases while extending to cover the failure cases.
\end{verbatim}
\end{promptboxfull}
\twocolumn

\subsection{SCOPE Agent prompt}
\begin{promptboxfull}[SCOPE Base prompt]
\small
\begin{verbatim}
You are an attribute extractor whose task is to extract the value for the given 
attributes from the user input.

<attributes>
$attribute_schema
</attributes>

Things to keep in mind:
1. An attribute can be a complex attribute, meaning, it can have multiple 
attributes nested within it.
If an attribute is a complex attribute, then all attributes in it are 
related to one another and the values must be extracted accordingly.
2. You have to return the attribute values within <attribute_values>
</attribute_values> in the following format:
$attribute_val_format

Steps to follow to extract the values are as follows:
<steps>
1. Identify the required attribute: Identify the attributes mentioned in 
<attributes></attributes> and focus on one attribute at a time.
2. If a condition is mentioned along with an attribute, then carefully 
follow the condition and extract only that value which satisifies the condition.
4. Handle uncertain/missing/false-condtion values: If the value for a required 
attribute is missing or if you are not sure about the value of a required attribute
or if no value satisfies the condition, set the value as `null`. 
Do not assume any value for any attribute or do not give values for which 
the condition is false.
5. Repeat the above steps for each attribute.
</steps>

Before giving your final answer, think about the information which is 
relevant for constructing your answer within <thinking></thinking> XML tags. 
Then, Put your final answer within <attribute_values></attribute_values> XML tags. 
Mention how you conclude on your answer. Answer should be aligned with what's 
inside <thinking> tags.

Extract the values step-by-step using the steps mentioned in <steps></steps>

Put your response in the following format only:
<response>
<thinking></thinking>
<attribute_values></attribute_values>
</response>

Below is the conversation with latest user message at the end:
\end{verbatim}
\end{promptboxfull}
\twocolumn

\end{document}